\documentclass[10pt,twocolumn,letterpaper,table]{article}
\usepackage{cvpr} 
\usepackage{graphicx}
\usepackage{amsmath}
\usepackage{amssymb}
\usepackage{booktabs}
\usepackage{multirow}
\usepackage{subcaption}
\usepackage{mathtools}
\usepackage{algorithm}
\usepackage{algpseudocode}
\usepackage[export]{adjustbox}
\usepackage[pagebackref,breaklinks,colorlinks]{hyperref}

\newcommand{\myrowcolour}{\rowcolor[gray]{0.925}}
\newcommand{\vecb}[1]{\boldsymbol{#1}}

\usepackage[capitalize]{cleveref}
\crefname{section}{Sec.}{Secs.}
\Crefname{section}{Section}{Sections}
\Crefname{table}{Table}{Tables}
\crefname{table}{Tab.}{Tabs.}

\begin{document}

\title{ProPLIKS: Probablistic 3D human body pose estimation}

\author{%
{Karthik Shetty$^{1,2}$} ~ {Annette Birkhold$^2$} ~ {Bernhard Egger$^1$} ~ {Srikrishna Jaganathan$^{1,2}$} ~ {Norbert Strobel$^2$}   \and ~  {Markus Kowarschik$^2$} ~  {Andreas Maier$^1$} ~  \\
\normalsize $^1$FAU~Erlangen-Nürnberg, Erlangen, Germany \quad
\normalsize $^2$Siemens~Healthineers~AG, Forchheim, Germany\\
{\tt\small karthik.shetty@fau.de}
}
\maketitle

\begin{abstract}
We present a novel approach for 3D  human pose estimation by employing probabilistic modeling. This approach leverages the advantages of normalizing flows in non-Euclidean geometries to address uncertain poses. Specifically, our method employs  normalizing flow tailored to the SO(3)  rotational group, incorporating a coupling mechanism based on the Möbius transformation. This enables the framework to accurately represent any distribution on SO(3), effectively addressing issues related to discontinuities. Additionally, we reinterpret the challenge of reconstructing 3D human figures from 2D pixel-aligned inputs as the task of mapping these inputs to a range of probable poses. This perspective acknowledges the intrinsic ambiguity of the task and facilitates a straightforward integration method for multi-view scenarios. The combination of these strategies showcases the effectiveness of probabilistic models in complex scenarios for human pose estimation techniques. Our approach notably surpasses existing methods in the field of pose estimation. We also validate our methodology on human pose estimation from RGB images as well as medical X-Ray datasets. 
\end{abstract}

\section{Introduction}
\label{sec:intro}
Reconstruction of 3D humans from 2D representations, such as images is an ill-posed  problem with inherent ambiguities. Traditionally, the norm  has been to produce a single, deterministic 3D prediction. However, it has been previously suggested that there are  benefits to considering a range of possible 3D poses from a given input~\cite{kolotouros2021probabilistic}. Tackling the challenge from a modern perspective, the use of deep neural networks has enabled the prediction of a probability distribution over 3D poses and shapes from image cues, addressing the difficulties of depth ambiguity, (self-)occlusion, and truncation. By using a probabilistic approach, we can improve upon the traditional methods by better estimating uncertainties, exploring various plausible 3D interpretations, and enhancing further applications such as integrating diverse data inputs (eg. multi-view) and improving model accuracy.

In the field of 3D human pose reconstruction, traditional methods  typically rely on deterministic approaches that generate a single 3D pose output. This preference is due to their simplicity and adequacy for various applications. However, the process of converting 3D information to a 2D image using a camera inherently leads to a loss of details. This information loss leads to a situation where the original 3D pose cannot be precisely recreated. Therefore, we propose that an effective 3D human pose estimator should be capable of presenting a range of possible 3D poses, each with its own probability. This paper emphasizes the use of probabilistic techniques while integrating the benefits of deterministic methods.

Previous probabilistic modeling techniques have utilized generative models such as VAE, normalizing flows, and diffusion models to create diverse pose hypotheses. Among these, normalizing flows stand out for their ability to efficiently sample and precisely estimate likelihoods in parametric model reconstruction. Unimodal distributions such as von Mises, Bingham, and matrix Fisher are particularly effective for modeling uncertainty when it is concentrated around a specific mode or direction. However, these distributions often require a slower sampling process, as they rely on rejection sampling methods.
The ProHMR~\cite{kolotouros2021probabilistic,zhou2019continuity} method uses  Gram-Schmidt projection for mapping rotations from 6D Euclidean space to the $\text{SO}(3)$ space, which disrupts the mapping characteristic of normalizing flows, leading to potential ambiguities and inaccuracies in modeling 3D rotational movements in human pose estimation. Meanwhile, HuManiFlow~\cite{sengupta2023humaniflow}, deriving from the ReLie~\cite{falorsi2019reparameterizing} framework, integrates normalizing flows for general Lie groups using Lie algebra in Euclidean space, but faces issues with discontinuous rotation representations, which compromise the accurate portrayal of inherently smooth rotational movements.

To address these challenges, we propose an approach that predicts a distribution of 3D poses that uniquely blends the concept of normalizing flows with the  Möbius transforms to accurately represent the 3D rotations on  the $\text{SO}(3)$ manifold. This technique enables probabilistic pose sampling, although it initially does not account for variations in camera perspectives or body shapes.  To address the latter, we model shape variations as a Gaussian distribution, interpreting the shape parameters as coefficients within a PCA shape space. To integrate these components effectively, we utilize PLIKS~\cite{shetty2023pliks}, which reformulates the SMPL model~\cite{smpl} into a series of linear equations that are solved using a least squares approach. This approach necessitates vertex predictions on the image plane, with uncertainties modeled through Laplacian Negative Log Likelihood (LNLL). Our approach is fully differentiable and integrates seamlessly into the training loop, enhancing model adaptability. Furthermore, we propose a strategy to deploy this trained model in multi-view systems without requiring additional retraining, thus broadening its potential applications.

Our paper's contributions can be summarized as follows:
\begin{itemize}
    \item We introduce a framework for probabilistic human body estimation that emphasizes generating multiple pose hypotheses and ensuring the alignment of the predicted poses with the images.
    \item  We contribute a Möbius transformation using basis transform to improve expressiveness in representing rotations within $\text{SO}(3)$ through normalizing flows.
    \item  We introduce a simplified isotropic $\text{SO}(3)$ Gaussian distribution for controlled mode representation and diverse sampling, providing an efficient alternative to the real isotropic Gaussian on $\text{SO}(3)$.
    \item  We establish that our method outperforms existing state-of-the-art techniques in various 3D human pose and shape benchmark evaluations.
    \item We demonstrate our method's adaptability by extending its application to X-ray imaging, potentially enhancing medical interventions with multi-view systems.
\end{itemize}

\section{Related Work}
The task of inferring human pose and shape from monocular images has been a subject of extensive study in recent years, incorporating both deterministic and probabilistic methodologies. Deterministic methods, as detailed in works such as~\cite{kanazawa2018end,kocabas2021pare,sun2021monocular}, primarily focus on deriving parameters from parametric body models like SMPL, utilizing the substantial statistical priors these models offer. Within this deterministic framework, there are two primary approaches: optimization and regression. Optimization methods, as exemplified in~\cite{bogo2016keep,fang2021reconstructing}, typically employ 2D keypoints derived from Deep Neural Networks (DNNs) and iteratively fit these keypoints to the SMPL model, though they are often subject to sensitivity regarding initial conditions and potential local optima. In contrast, regression methods \cite{kanazawa2018end,pavlakos2018learning,tung2017self,guler2019holopose} directly predict pose and shape parameters through DNNs but require significant amounts of training data. A blend of these these approaches is seen in SPIN~\cite{kolotouros2019learning}, which merges optimization and regression to enhance mesh accuracy supervision. This approach, while facing challenges in mesh-image alignment, has been further refined in subsequent studies such as~\cite{li2021hybrik,iqbal2021kama,shetty2023pliks}. Model-free strategies diverge from these methods by directly regressing vertices through intermediate representations, thus forging a correlation between the model and the input image~\cite{moon2020i2l,lin2021end,guler2018densepose}.

Complementing these deterministic models are probabilistic or multiple hypothesis methods, which rely on a spectrum of modeling techniques. Earlier iterations of these methods often employed heuristic strategies to sample a variety of plausible 3D joint configurations~\cite{moreno20173d,simo2012single}. Recent developments have introduced methodologies based on mixture density networks~\cite{li2019generating,oikarinen2021graphmdn}, and conditional variational autoencoders~\cite{sohn2015learning,sharma2019monocular}. Moreover, diffusion-based methods for 3D keypoints have been proposed~\cite{gong2023diffpose,holmquist2023diffpose}, and even in adjacent fields such as motion hypothesis, diffusion models have shown efficacy in parametric modeling~\cite{tevet2022human}. The application of normalizing flows in understanding more intricate distributions over parametric models is explored in~\cite{kolotouros2021probabilistic}, while~\cite{sengupta2023humaniflow} utilizes~\cite{falorsi2019reparameterizing} for flows on the Lie group via the Lie algebra $\mathfrak{so}(3)$.

In the specific area of probabilistic distributions over $\text{SO}(3)$ for rotation regression, diverse approaches have been adopted. A method using a mixture of von Mises distributions over Euler angles is presented in~\cite{prokudin2018deep}. The application of matrix Fisher distribution for deep rotation regression, which has also been adapted for human pose estimation, is discussed in~\cite{mohlin2020probabilistic,sengupta2021hierarchical}. Additionally, a recent study~\cite{liu2023delving} has introduced a general concept of Möbius coupling for rotation regression  over $\text{SO}(3)$. Unlike ~\cite{liu2023delving}, we implement a basis transform that allows us to extend transformations to the entire circle, rather than just the semi-circle as previously proposed. ~\cite{dunkel2024normalizing} suggests utilizing the flow from ~\cite{liu2023delving} to generate multiple pose hypotheses, although it does not incorporate camera and translation aspects.

\section{Methodology}
In this section, we introduce our network architecture  as illustrated in Fig.~\ref{fig-arc}, which features a  normalizing flow module designed specifically for estimating rotations within the $\text{SO}(3)$ manifold. We start with a discussion of the SMPL model, including its forward kinematics process. We then move on to the Möbius transformation, which originates from the circular-circular regression model, followed by an exploration of the Möbius coupling layer that adapts this transformation for the $\text{SO}(3)$ manifold. Subsequently, we offer a brief overview of the base isotropic Gaussian distribution.  We conclude this section with a detailed description of the entire architecture, which outlines the methodology for mesh recovery from RGB or X-Ray images using conditional normalizing flows.

\subsection{Human Body Parametrization}
For characterizing the human body, we can utilize SMPL~\cite{smpl}, BOSS~\cite{boss},  OSSO~\cite{osso} or any other paramteric model, depending on the specific use-case.  SMPL is adopted for estimating human models from RGB images, while BOSS is tailored for visualizing human skeletons from X-Ray images. Given that both models share an identical structure, we provide a unified architectural overview. The model is conceptualized as a statistical parametric function, symbolized as $\mathcal{M}(\vecb{\beta}, \vecb{\theta}; \vecb{\Phi})$, which produces a corresponding surface mesh.

The shape attributes $\vecb{\beta}$, employ principal component analysis to linearly project the basis $\mathbf{B}$ from $\mathbb{R}^{|\vecb{\beta}|}$ to $\mathbb{R}^{3 N}$, where $N$ denotes the number of vertices. This projection captures variations from the mean mesh, $\vecb{\Bar{x}}_m$, as expressed by the equation $\vecb{x} = \vecb{\Bar{x}}_m + \vecb{\beta}\mathbf{B}$. The pose of the model is characterized by a kinematic chain, consisting of an array of relative rotation vectors $\vecb{\theta}=\left[\vecb{\theta}_{1}, \ldots, \vecb{\theta}_{K}\right] \in \mathbb{R}^{K \times 3}$, represented via the axis-angle representation for the $K$ joints. 

Additional parameters contained within $\mathbf{\Phi}$ are integral to the deformation mechanism of the model. These include the joint regressor $\mathcal{J}\in\mathbb{R}^{K\times 3N}$, blend coefficients $\mathcal{W}\in\mathbb{R}^{K\times N}$, and pose-specific shape transformations, represented as $B_{P}(\vecb{\theta})$. It's important to point out that the BOSS model does not account for pose-specific deformation because the models were trained using a static CT dataset, and bones are rigid structures. The resultant body mesh is constructed from an initial template by applying the relevant forward kinematics, based on the relative rotations, $\vecb{\theta}$, and shape alterations, $\vecb{\beta}$.

\subsection{Möbius Flow}
The Möbius transformation is a bijective mapping that transforms the complex plane $\mathbb{C}$, by preserving the structure of circles and lines. Specifically, it can map the unit circle either to itself or to a straight line. Based on prior work that leveraged Möbius transformations for a circular-circular regression model~\cite{kato2008circular}, we integrate this approach to form the basis of our Möbius coupling layer tailored for the $\text{SO}(3)$ manifold:
\begin{equation}
\label{eq:cc}
    \hat{c} = \frac{c + \omega}{\omega^*  c + 1}.
\end{equation}
In the equation above, $c \in \mathbb{C}$  resides on the unit circle, while $\omega \in \mathbb{C}$ is a distinct complex parameter ensuring $|\omega| \neq 1$. Intuitively, $\omega$ can be perceived as the parameter pulling the circle's points toward $\omega/|\omega|$, with the concentration around $\omega/|\omega|$ intensifying as $|\omega|$ grows~\cite{kato2008circular}. It's worth noting that the expression in Eq.~\ref{eq:cc} is equivalent to (Eq.8) from~\cite{rezende2020normalizing} which extends to a unit $D$-dimensional sphere $\mathbb{S}^{D}$ in $\mathbb{R}^{D+1}$  as illustrated in~\cite{kato2015m}. A fundamental observation from~\cite{rezende2020normalizing,kato2015m} is that Möbius transformations exhibit group properties under function composition. This implies that merely stacking multiple transformations doesn't enhance their expressiveness. To address this limitation, a convex combination of numerous such transformations is employed. Yet, for this strategy to be effective, the transformations must exhibit monotonic behavior~\cite{rezende2020normalizing}. 

\subsubsection{Möbius Coupling}
For modeling $\text{SO}(3)$, we adopt the strategy of coupling layers as outlined in~\cite{dinh2016density}. Rotation matrices $\text{R} \in \text{SO}(3)$ can be depicted using two orthonormal vectors passing through the origin. The third vector can be deduced through the Gram-Schmidt procedure by taking the cross product. In this context, let $\vecb{u}_1 \in \mathbb{R}^3$ and $\vecb{u}_2 \in \mathbb{R}^3$  be the orthonormal vectors supplied to the Möbius coupling layer. In the initial coupling layer, we keep $\vecb{u}_1$ intact, and apply the Möbius transformation to $\vecb{u}_2$. 

We start with vectors $\vecb{u}_1$ and $\vecb{u}_2$ to create the third orthonormal vector $\vecb{u}_3 = \vecb{u}_2 \times \vecb{u}_1$. Our goal is to convert $\vecb{u}_2$ into $\vecb{\bar{u}}_2\in \mathbb{R}^3$ in a way that is reversible and keeps $\vecb{\bar{u}}_2$ within the plane formed by $\vecb{u}_3$ and $\vecb{u}_2$. Since $\vecb{u}_3$ and $\vecb{u}_2$ are orthonormal, any vector $\vecb{v} \in \mathbb{R}^3$ can be projected onto them to find its coordinates in this basis. Thus, $\vecb{v}$ can be represented as a complex number $\vecb{\Tilde{v}} = \langle \vecb{v}, \vecb{u}_2 \rangle + i \langle \vecb{v}, \vecb{u}_3 \rangle$, where $\langle \cdot, \cdot \rangle$ denotes the dot product. Transforming the basis for $\vecb{u}_2$ results in $\vecb{\Tilde{u}}_2 \in \mathbb{C}$, which is equivalent to a complex number $1 + i0$, acting as a fixed reference.

\begin{figure}[!b]
\begin{center}
\includegraphics[width=1.0\linewidth,]{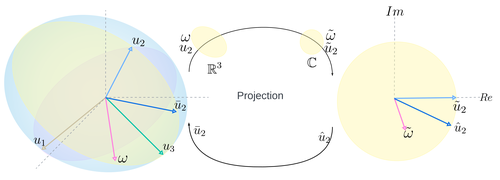}
\end{center}
\vspace{-15pt}
\caption{In the Möbius layer, $\vecb{u}_1$ acts as the identity vector, which is conditioned by a neural network to derive $\vecb{\omega}$. The vector $\vecb{u}_2$ is subjected to the Möbius transformation. The third vector, $\vecb{u}_3$, is generated from the cross product of $\vecb{u}_2$ and $\vecb{u}_1$. We project $\vecb{u}_2$ and $\vecb{\omega}$ onto the complex plane (indicated in yellow) using the basis vectors $\vecb{u}_2$ and $\vecb{u}_3$. The Möbius transformation is then applied to produce the transformed vector $\vecb{\hat{u}}_2$, which is subsequently re-projected onto the 3D space to yield $\vecb{\bar{u}}_2$.}
\label{fig-2}
\end{figure}

\begin{figure*}[!ht]
\centering
\includegraphics[width=1.0\linewidth,keepaspectratio]{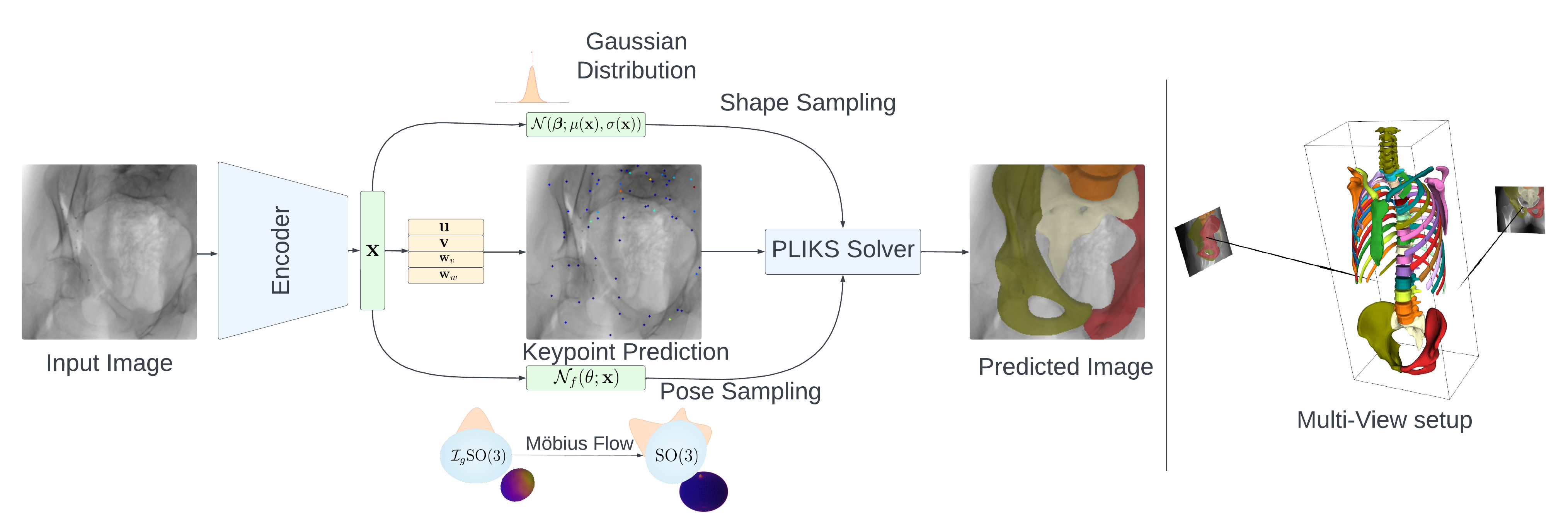}
\vspace{-15pt}
  \caption{\textbf{Overview of the proposed framework:} Starting with an input image, the backbone of the network generates a feature vector $\mathbf{x}$ that is used for sampling poses via normalizing flow and shapes through Gaussian sampling. Concurrently, it predicts a sparse set of keypoints on the image plane, each associated with a measure of uncertainty. These sampled poses and shapes are then input into the PLIKS solver, which optimizes the parameters for pose, shape, and translation based on the predicted keypoints. Additionally, the solver has the capability to process multiple images simultaneously, integrating individual predictions from each image sample.}
\label{fig-arc}
\end{figure*}

Consider a vector $\vecb{\omega} \in \mathbb{R}^3$, conditioned on $\vecb{u}_1$ and possibly other features using neural networks. We apply the same basis transformation to $\vecb{\omega}$ as described above, resulting in $\vecb{\Tilde{\omega}} \in \mathbb{C}$, ensuring that $|\vecb{\Tilde{\omega}}| \neq 1$. By applying the Möbius transformation from Eq.~\ref{eq:cc}, we derive a new vector $\vecb{\hat{u}}_2 \in \mathbb{C}$. Finally, we project $\vecb{\hat{u}}_2$ back onto the original basis vectors $\vecb{u}_3$ and $\vecb{u}_2$ within $\mathbb{R}^3$. This is visualized in Fig.~\ref{fig-2}

As previously discussed, simply combining multiple transformations does not enhance expressivity. Instead, we use a convex combination of Möbius transformations as illustrated in~\cite{rezende2020normalizing}. This method involves applying multiple Möbius transformations to the input vector $\vecb{\Tilde{u}}_2$, resulting in a set of $k$ transformed vectors $\vecb{\hat{u}}_2^k$, each modified by a distinct $\vecb{\Tilde{\omega}}^k$. Subsequently, we compute the weighted sum $\theta = \sum_k w_k \theta_k$, where $w_k \in \mathbb{R}$ represents a weight, derived from a neural network conditioned in a manner similar to $\vecb{\omega}$, with the constraint that $\sum_k w_k = 1$. The $\theta_k$ values represent the angles of the transformed vectors, defined as  $\text{arg}(\vecb{\hat{u}}_2^k)$. The composite $\theta$, which is a unit-length complex vector, is then projected back onto the original basis vectors.  To ensure that this transformation is invertible, it is crucial to maintain that $|\omega| < 1$.

The primary motivation for changing the basis is to establish a consistent reference derived from the input, eliminating the need to have a predefined one. This inherent reference ensures the function's monotonicity, enabling its numerical inversion through bisection search~\cite{rezende2020normalizing}. The challenges associated with setting an explicit reference, or choosing alternatives methods such as non-compact projection~\cite{rezende2020normalizing} or circular splines~\cite{rezende2020normalizing}, are analogous. Consider all methods transform an angle $\theta$ into a new angle $\hat{\theta} = f(\theta)$ via a monotonic function $f : [0, 2\pi] \to [0, 2\pi]$. The characteristics of $f$ around $0$ and $2\pi$ isn't guaranteed to be mirrored. If, during training, the input angle approaches $2\pi$ from values near $0$ (essentially becoming negative), the transformation dictated by $f$ could vary significantly, leading to pronounced downstream changes in the network. This variability introduces training instability. As a remedy, we adopt a basis change, enhancing network stability during training. Contrary to the approach suggested in~\cite{liu2023delving}, which enforces $|\omega| < \frac{\sqrt{2}}{2}$ to guarantee invertibility and consequently transforms to only half of the circle, our method can map the full circle, thereby enhancing its expressiveness. Details on the inverse and Jacobian can be found in the supplementary materials.

\subsubsection{Rotation Transformation}
To complement the Möbius coupling layer, we incorporate a rotation transformation. Given a conditional vector,  a network based transformation produces a 9D vector and projects it onto $\text{SO}(3)$ using symmetric orthogonalization~\cite{levinson2020analysis}. This projection is represented as $U V^T \text{diag}(1, 1, \det(UV^T))$
where $U$ and $V^T$ are orthogonal matrices derived from the SVD. As this process solely enacts a rotation, it preserves volume element. It has been demonstrated that the combination of rotational transformations alongside Möbius transforms results in better expressiveness~\cite{liu2023delving}.

\subsubsection{Isotropic Gaussian Distribution}
Normalizing flow networks, when applied in the context of human pose estimation, typically start by transforming samples from a simple known probability distribution, often a Gaussian~\cite{kolotouros2021probabilistic}. The Gaussian distribution inherently has a mode representing the peak probability, and the samples drawn from it inherently exhibit the characteristic bell curve. Given that our model anticipates inputs from $\text{SO}(3)$, we adopt a simplified approach inspired by the projected isotropic normal distribution to handle isotropic Gaussian distribution. This approach facilitates feasible inference over the unit sphere~\cite{hernandez2017general}. Every rotation matrix in $\text{SO}(3)$ can be represented by a quaternion, which resides on the  $3$-sphere in $\mathbb{R}^4$. However, directly sampling from a  $3$-sphere  presents challenges due to the double-cover nature of quaternions. To mitigate this, we can adjust the concentration factor of our distribution to control its variance, ensuring that the samples primarily fall within one hemisphere. While a true normal distribution on the rotation group $\text{SO}(3)$~\cite{nikolayev1997normal,leach2022denoising} overcomes the drawbacks of quaternions such as double covering, our simplified method suffices for our specific objective ensuring that the model remains practical while capturing the necessary rotational behaviors.

\subsection{Stochastic Mesh Reconstruction}
Our mesh recovery framework generally aligns with the architecture of ProHMR~\cite{kolotouros2021probabilistic}. We aim to determine a distribution of potential shapes and poses of a person, conditioned on the input image. Consequently, our network produces a conditional probability distribution $p_{\Theta \mid I}(\vecb{\theta},\vecb{\beta} \mid I)$, representing the likelihood of various shapes and poses for a given image. Utilizing CNN as the backbone neural network, we extract the features $\mathbf{x}$ from the provided image. We employ the previously discussed Möbius flow as our normalizing flow network to represent poses in $\text{SO}(3)$. For shape representation, a conditional Gaussian distribution is utilized~\cite{sengupta2021hierarchical}. Both are influenced by the conditional vector $\mathbf{x}$.

A challenge arises when establishing a probability distribution on shape, particularly concerning the estimation of camera parameters, notably the weak camera parameter. To address this, we incorporate PLIKS~\cite{shetty2023pliks}, capable of determining the translation using an initial pose estimate $\vecb{\theta}$ and a sparse set of pixel-aligned vertices $\vecb{i} $. This is achieved by reframing SMPL as a least squares problem and linearizing the SMPL model, this issue is effectively resolved. 

\begin{equation}\label{eq_argmin}
    \underset{ \mathbf{{\Delta R}},\vecb{\beta},\vecb{t}}{\mathrm{argmin}}|| \vecb{w}^{\Tilde{k}}\Big(\vecb{i}^{\Tilde{k}} - \mathbf{\hat{K}} ( \mathbf{{\Delta R}} {\vecb{x}_r^{\Tilde{k}}}   + \vecb{\beta} {\mathbf{B}_r^{\Tilde{k}}}  +  \vecb{t}\mathbf{W}_r^{\Tilde{k}})\Big)||_2 {+} \omega_\beta||\vecb{\beta}||_2.
\end{equation}

Here, $\vecb{i} \in \mathbb{R}^{N \times 2}$, represents the sparse set of keypoints predicted by the network, positioned on the image plane and specified in $(u, v)$ coordinates. $\vecb{\hat{K}} {\in} \mathbb{R}^{3 \times 4}$ matrix denotes the intrinsic parameters of the cropped image. While, $\mathbf{\Delta R}_k$ is defined as the additional rotation required to achieve an optimal solution, as outlined in~\cite{shetty2023pliks}. For further definitions of variables, please refer to~\cite{shetty2023pliks}.

Additionally, we calculate a weighting factor $\vecb{w} = {\vecb{w}_v * \vecb{w}_w} \in \mathbb{R}^{N}$. Here, $\vecb{w}_w$ represents the unsupervised weighting factor as proposed in ~\cite{shetty2023pliks}. Each predicted keypoint is treated as a stochastic variable, influenced by the probability density function of a 2D Laplacian. These, individual keypoint predictions can be characterized by $\{\vecb{i}, \vecb{w}_v\}$, symbolizing the mean and variance, respectively. In this representation, the mean signifies the expected landmark position, while the variance offers an estimation of prediction uncertainty. For instances with known ground truth $\vecb{\hat{i}}$, the mean can be utilized directly, however, the variance is determined in an unsupervised manner.  This is facilitated by training the network using a Laplacian negative log likelihood loss function as follows
\begin{equation*}
\mathcal{L}_{lnll} =  \log(2\vecb{w}) + \frac{||\vecb{\hat{i}} -  \vecb{i}||_1}{\vecb{w}}
\end{equation*}

$\vecb{w}_w$ is the unsupervised weigthing as propposed in ~\cite{shetty2023pliks}.

From the derived samples, we preliminary compute the rotation influencing the average shape ${\vecb{x}_r^{\Tilde{k}}}$ and the shape basis vector ${\mathbf{B}_r^{\Tilde{k}}}$. Contrary to PLIKS, our method doesn't configure the mean shape to zero  $\omega_\beta||\vecb{\beta}||_2$, but rather to the estimates obtained via the stochastic approach. This strategy eliminates the necessity for stabilization phases, enabling a seamless end-to-end training of the network.  Utilizing a linear solver, we observe that many pose hypotheses are effectively countered, as the solver aims to optimally align with the pixel-aligned vertices. To mitigate this effect, we apply a regularizer specifically on pose by setting $\mathbf{{\Delta R}}_k = \gamma \mathbf{I}$ for $k = 1, \ldots, K$. This ensures that outputs from the solver are closer to the sampled poses derived from the normalizing flow.

\section{Experiments}
The experiments are divided between RGB and X-Ray images. For RGB images, we exclusively trained the model on synthetic datasets, namely BEDLAM~\cite{bedlam} and AGORA\cite{agora}. The assessment of this model is performed on the Human3.6M~\cite{h36m} and 3DPW~\cite{3dpw} datasets, with an emphasis on outcomes from networks that also incorporate real datasets into their training.  Further details about these datasets and their processing methods along with the X-ray experiments are provided in the supplementary materials.

\subsection{RGB Images}
In this section, we assess the precision of our model by comparing it with existing deterministic and probabilistic approaches. Our findings indicate that our method, even though it solely relies on synthetic data, outperforms previous methods. We also demonstrate multiple hypotheses of our approach, showcasing its enhanced performance over prior techniques. For consistency with established methodologies, we employ the LSP regressor~\cite{kolotouros2019learning} to derive the 14 joint locations for the Human3.6M and 3DPW datasets. We evaluate our method using the mean per joint position error (MPJPE), the procrustes aligned mean per joint position error (PA-MPJPE), and errors in 2D keypoints (2DKP).

\begin{table*}[]
\centering
\small	
\begin{tabular}{l|c|cc|cc}
\multicolumn{1}{c|}{\multirow{2}{*}{Method}} & \multirow{2}{*}{Data} & \multicolumn{2}{c|}{3DPW}                       & \multicolumn{2}{c}{Human3.6M}                  \\ \cline{3-6} 
\multicolumn{1}{c|}{}                        &                          & PA-MPJPE~$\downarrow$                           & MPJPE~$\downarrow$       & PA-MPJPE~$\downarrow$                         & MPJPE~$\downarrow$      \\ \hline
HMR$\ddag$~\cite{kanazawa2018end}                               & R                        & \multicolumn{1}{c|}{81.3 /   -}   & 130 /  -    & \multicolumn{1}{c|}{56.8/  -}    & 88.0/  -    \\ \myrowcolour
SPIN$\ddag$~\cite{kolotouros2019learning}                             & R                        & \multicolumn{1}{c|}{59.2 /  -}    & 96.9 /  -   & \multicolumn{1}{c|}{41.1/  -}    & - / -       \\
PARE$\ddag$$^\dagger$~\cite{kocabas2021pare}                             & R                        & \multicolumn{1}{c|}{46.5 /  -}    & 74.5 /  -   & \multicolumn{1}{c|}{- / -}       & - / -       \\ \myrowcolour
CLIFF$\ddag$$^\dagger$~\cite{li2022cliff}                           & R                        & \multicolumn{1}{c|}{43.0 /  -}    & 69.0 /  -   & \multicolumn{1}{c|}{\textbf{32.7}/  -}    & 47.1/  -    \\
PLIKS$\ddag$$^\dagger$~\cite{shetty2023pliks}                           & R                        & \multicolumn{1}{c|}{\textbf{38.5} /  -}    & \textbf{60.5} /  -   & \multicolumn{1}{c|}{34.5/  -}    & \textbf{47.0}/  -    \\ \myrowcolour
BEDLAM-HMR~\cite{bedlam}                   & S                        & \multicolumn{1}{c|}{47.6 /  -}    & 79.0 /  -   & \multicolumn{1}{c|}{51.7/  -}    & 81.6/  -    \\ 
BEDLAM-CLIFF~\cite{bedlam}                   & S                        & \multicolumn{1}{c|}{46.6 /  -}    & 72.0 /  -   & \multicolumn{1}{c|}{50.9/  -}    & 70.9/  -    \\ \myrowcolour \hline 
ProHMR$\ddag$~\cite{kolotouros2021probabilistic}                               & R                        & \multicolumn{1}{c|}{59.8 / 48.2}  & 97.0 / 81.5 & \multicolumn{1}{c|}{\textbf{41.2} /   -}  & - / -       \\
HuManiFlow~\cite{sengupta2023humaniflow}                           & S                        & \multicolumn{1}{c|}{53.4 / 39.9}  & 83.9 / 65.1 & \multicolumn{1}{c|}{- / -}       & - / -       \\  \myrowcolour
Proposed$_{\text{w/o solver}}$                                     & S                        & \multicolumn{1}{c|}{47.8 / 37.6} & 71.6 / 57.6 & \multicolumn{1}{c|}{51.1 / 41.8} & 79.2 / 67.4\\ 
    Proposed                          & S                        & \multicolumn{1}{c|}{\textbf{45.4} / \textbf{34.5}} & \textbf{69.7} / \textbf{55.8}& \multicolumn{1}{c|}{44.4 / \textbf{35.1}} & \textbf{73.6} / \textbf{59.6}
\end{tabular}
\caption{\label{tab:main}Benchmark of state-of-the-art models on 3DPW and Human3.6M datasets. All results are in mm.  Here, S represents synthetic and R real training dataset, while $\dagger$, $\ddag$ denotes models trained on 3DPW and Human3.6M dataset respectively. Deterministic approaches are on the top, followed by probabilistic approaches. For probabilistic methods, values are displayed as mode / minimum for 3D  based on a sample size $n = 100$.}
\end{table*}

\subsubsection{Comparison with the State-of-the-art}
Our approach is evaluated against prior techniques on the Human3.6M~\cite{h36m} and 3DPW~\cite{3dpw} datasets. Table~\ref{tab:main} presents quantitative comparisons with prior methods. For our comparison, we select the optimal results from other studies, specifically those models trained on the 3DPW dataset. We detail the performance of our model both with and without the use of the PLIKS solver. In instances where the solver is not employed, we utilize a weak perspective camera model with an extended focal length. To mitigate the distortion effects caused by camera warping during the cropping process, we incorporate an implicit camera rotation in the post-processing phase~\cite{mpiinf}.

When sampled from a Gaussian distribution, the mode of our network surpasses the comparable network BEDLAM-CLIFF~\cite{bedlam} regarding both model backbone and training dataset, as evident by the performance in the evaluated metrics. Notably, we find that sampling from the distribution yields the lowest MPJPE on the 3DPW dataset, irrespective of whether the solver is used or not.Although our results do not surpass those of deterministic methods, it's important to highlight that our models were trained exclusively on synthetic images, resulting in a domain gap.

ProPLIKS delivers more precise distributions and achieves lower minimum sample metrics compared to other probabilistic methods. Although the model performs better without the solver in terms of MPJPE and PA-MPJPE, the 2D keypoint (2DKP) accuracy significantly improves with the solver, leading to more meaningful sample outcomes. This is further discussed in the following section. In Fig.~\ref{fig:final}, we present qualitative outcomes of our method, highlighting that the model generates samples that not only closely match 2D observations but also exhibit a wide variety in 3D predictions.

\subsection{X-Ray Images}

In the case of X-Ray images, our training utilized the CTLymph~\cite{lymph} and AutoPet~\cite{autopet} datasets. For validation, we utilized the QIN-HeadNeck~\cite{qin} and Visceral~\cite{visceral} datasets.  We created Digitally Reconstructed Radiograph (DRR) images from CT scans as synthetic inputs for the network.
We make use of TotalSegmentator~\cite{wasserthal2023totalsegmentator} to segment the bone structures and then transform these segmentations into surface meshes. Traditional iterative optimization techniques are then applied to fit the BOSS model to these surface scans. The parameters obtained from this process are used for both training and validation purposes. Reconstruction on single view images are provided in the supplemental material.

\begin{table}[b]
\centering
\scriptsize
\begin{tabular}{c|cc|cc|cc}
                            & \multicolumn{2}{c|}{MPJPE~~$\downarrow$~(mm)}     & \multicolumn{2}{c|}{MPJPE (abs)$\downarrow$~(mm)} & \multicolumn{2}{c}{2DKP$\downarrow$~(mm)}    \\ \cline{2-7} 
                            & N=1                  & N=2                  & N=1                     & N=2                     & \multicolumn{1}{c}{N=1} & \multicolumn{1}{c}{N=2} \\ \hline
Pelvis                      & 11.10  & 8.92                       & 69.01   & 9.54                        & 7.34                       & 7.67  \\ \myrowcolour 
Lumbar                      & 22.09  & 16.44                       & 59.33   & 12.96                        & 11.82                       & 12.75  \\ 
Thorax                      & 33.66  & 26.59                       & 75.02   & 20.14                        & 9.37                      & 11.27  \\ 
\end{tabular}%
\caption{\label{tab:xray2} Assessment on a multi-view X-Ray dataset is presented. Here, abs refers absolute distance, while N=1 refers to single view, whereas N=2 refers to multi-view, where the cameras are at least $30^\circ$ apart.}
\end{table}

\begin{table*}[!ht]
\footnotesize
\centering
\begin{tabular}{l|l|c|c|c}
Changes                               & Method                  & PA-MPJPE~$\downarrow$ (mm) & MPJPE~$\downarrow$ (mm) & 2DKP~$\downarrow$ (px) \\ \hline
\multirow{3}{*}{Flow$^\dag$}                 & \textbf{Proposed}       & 47.8 / 37.6                & 71.6 / 57.6             & 6.3 / 8.3              \\
& Möbius~\cite{liu2023delving}          & 50.0 / 39.9                & 73.1 / 60.3             & 6.7 / 8.8              \\
& GLOW~\cite{kolotouros2021probabilistic}            & 49.1 / 40.2                & 73.2 / 60.8             & 6.5 / 8.2              \\ \hline
\multirow{2}{*}{Distribution$^\dag$}         & \textbf{Proposed IGSO3}& 47.8 / 37.6                & 71.6 / 57.6             & 6.3 / 8.3              \\
                                      & IGSO3                   & 47.9 / 38.1                & 71.5 / 57.9             & 6.3 / 8.4              \\ \hline
\multirow{2}{*}{Concentration} & \textbf{\boldmath$\kappa=2$}     & 45.4 / 34.5                & 69.7/ 55.8              & 5.8 / 6.5 \\
                                      & $\kappa=3$             & 45.7 / 39.5                & 70.1 / 60.3             & 5.6 / 5.9              \\ \hline
Solver gamma                          & \textbf{\boldmath$\gamma=0.5$}           & 45.4 / 34.5                & 69.7/ 55.8              & 5.8 / 6.5               \\
                                      & $\gamma=0.1$   & 45.4 / 37.8                & 68.6/ 58.3              & 5.7 / 5.9               \\ \hline
\multirow{2}{*}{Dataset B+A} & \textbf{Proposed}       & 45.4 / 34.5                & 69.7/ 55.8              & 5.8 / 6.5               \\
                                      & BEDLAM-CLIFF~\cite{bedlam}            & 47.4 / -                   & 73.0 / -                & - / -                  \\
\multirow{2}{*}{Dataset A}        & \textbf{Proposed}       & 48.7 / 41.2                & 72.7 / 63.0             & 5.9 / 6.9              \\
                                      & BEDLAM-CLIFF~\cite{bedlam}            & 54.0 / -                   & 88.0 / -                & - / -                 
\end{tabular}\caption{Exploring different configurations in the model design ablation study. In this context, $\dag$ indicates the absence of the solver, aligning with the HMR approach~\cite{kanazawa2018end,bedlam}. The symbol B denotes the Bedlam dataset~\cite{bedlam}, while A stands for the Agora dataset~\cite{agora}. Best method is highlighted in bold.\label{Table:abl}}. 

\end{table*}

\subsubsection{Multi-View}
X-Ray imaging typically involves acquiring images with established calibration parameters. We demonstrate the application of this in a multi-view context. For this purpose, we consider having a set of calibrated images, each with known intrinsic parameters $\mathbf{K}_j$ and extrinsic parameters $\mathbf{[R^e|t^e]}_j$. In a multi-view setup using our solver, it's imperative that the pose, shape, and translation parameters remain consistent across all views. We start by averaging the poses from different views for each sample set, employing SVD to derive an initial approximation of poses. Subsequently, we modify the minization function, as follows
\begin{equation}
|| \sum_j\vecb{w}^{\Tilde{k}} \mathbf{R}^e_j\Big(\vecb{i}^{\Tilde{k}} - \mathbf{\hat{K}} ( \mathbf{{\Delta R}} {\vecb{x}_r^{\Tilde{k}}}   + \vecb{\beta} {\mathbf{B}_r^{\Tilde{k}}}  +  \vecb{t}\mathbf{W}_r^{\Tilde{k}})\Big) + \mathbf{t}^e_j||_2,
\end{equation}
keeping in mind using similar approximations to those used in PLIKS~\cite{shetty2023pliks}. 
Table~\ref{tab:xray2} reveals a substantial decrease in absolute error when employing a multiview system. It is noteworthy that this improvement is achieved without training a new network; rather, the same network initially trained on single-view data is utilized.

\subsection{Ablations}
In this context, we evaluate the individual components of our approach as shown in Table ~\ref{Table:abl}. Our evaluations exclusively use the 3DPW dataset for testing, while the AGORA and BEDLAM datasets are employed for training. Initially, we assess the performance of our designed flow against existing methodologies. Given that ProHMR~\cite{kolotouros2021probabilistic} was developed on a different dataset, we re-implemented it using our network architecture and hyper-parameters for a fair comparison, with detailed values available in the supplementary materials. Additionally, we examine the normalizing flow introduced by~\cite{liu2023delving}. To assess variations in the flow architecture, we conduct our evaluations without employing solvers, with the framework similar to a conventional HMR network~\cite{kanazawa2018end}.
Our findings indicate that our proposed flow outperforms others across all evaluated metrics.

\begin{figure}[hb!]
  \centering
  \begin{subfigure}[b]{0.24\linewidth}
    \includegraphics[width=\linewidth]{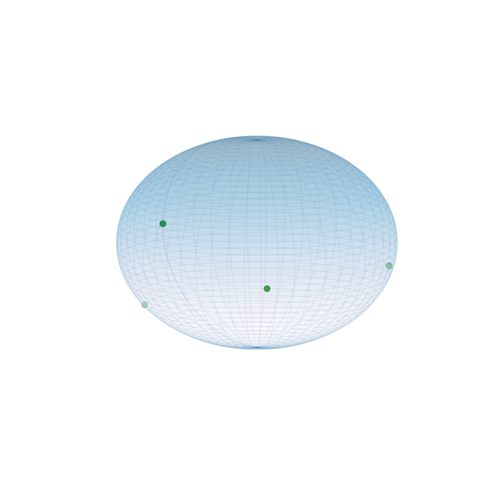}
    \caption{GT}
    \label{fig:figure1}
  \end{subfigure}
  \begin{subfigure}[b]{0.24\linewidth}
    \includegraphics[width=\linewidth]{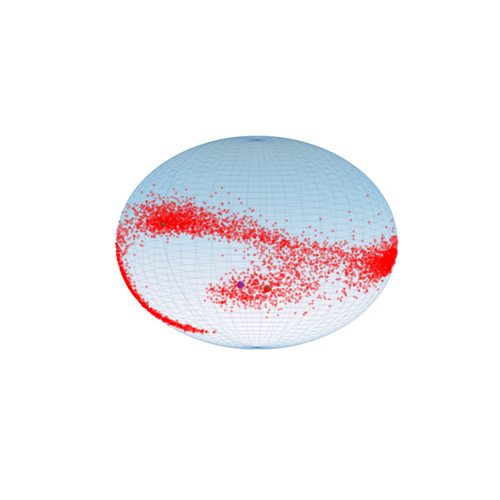}
    \caption{GLOW~\cite{kolotouros2021probabilistic}}
    \label{fig:figure2}
  \end{subfigure}
  \begin{subfigure}[b]{0.24\linewidth}
    \includegraphics[width=\linewidth]{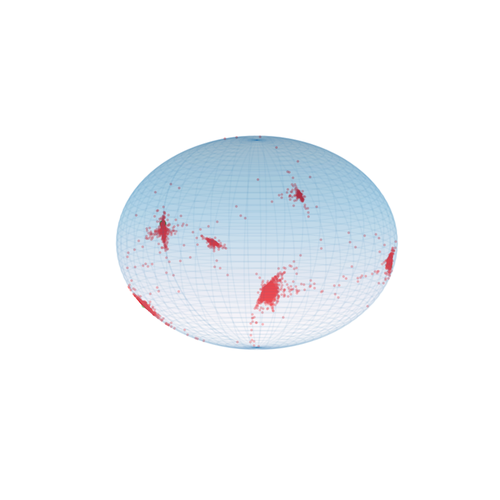}
    \caption{ReLie~\cite{sengupta2023humaniflow}}
    \label{fig:figure3}
  \end{subfigure}
  \begin{subfigure}[b]{0.24\linewidth}
    \includegraphics[width=\linewidth]{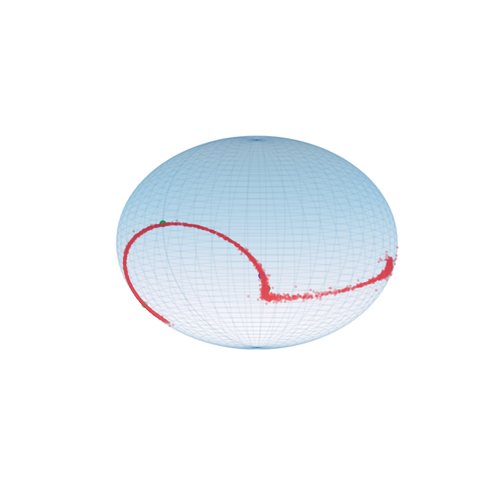}
    \caption{Proposed}
  \end{subfigure}
  
  \caption{Variations in flow sampling. Here, green represents the rotation distribution that needs to be learned, red indicates the samples obtained after training the flow model, and blue represents the mode of the distribution.}
  \label{fig:four_figures}
\end{figure}

Further, we investigate how well the proposed Isotropic Gaussian distribution aligns with an actual distribution~\cite{nikolayev1997normal,leach2022denoising}. Despite observing comparable performance, we favor our approach for its significantly lower memory requirements during sampling. Theoretically, adjusting the concentration factor of the projected normal distribution alters its variance; a higher concentration results in a denser distribution, which is vital for ensuring that the generated samples are confined to a single hemisphere. Reducing the concentration diminishes network effectiveness.

We also explore the impact of solver weighting $\mathbf{{\Delta R}}_k = \gamma \mathbf{I}$, which diminishes the solver's influence in favor of the network's predicted poses. This adjustment leads to an increase in 2D pixel errors and greater diversity in 3D representations. Notably, the 2D errors remain relatively low across a broad spectrum of 3D variations, especially when compared to scenarios excluding the PLIKS solver. In this analysis, it is evident that despite a broad array of 3D variations, the 2D error is kept to a minimum, especially when contrasted with scenarios where the PLIKS solver is not utilized.

\begin{figure*}[!h]
	\centering
 \includegraphics[width=0.8\linewidth,keepaspectratio]{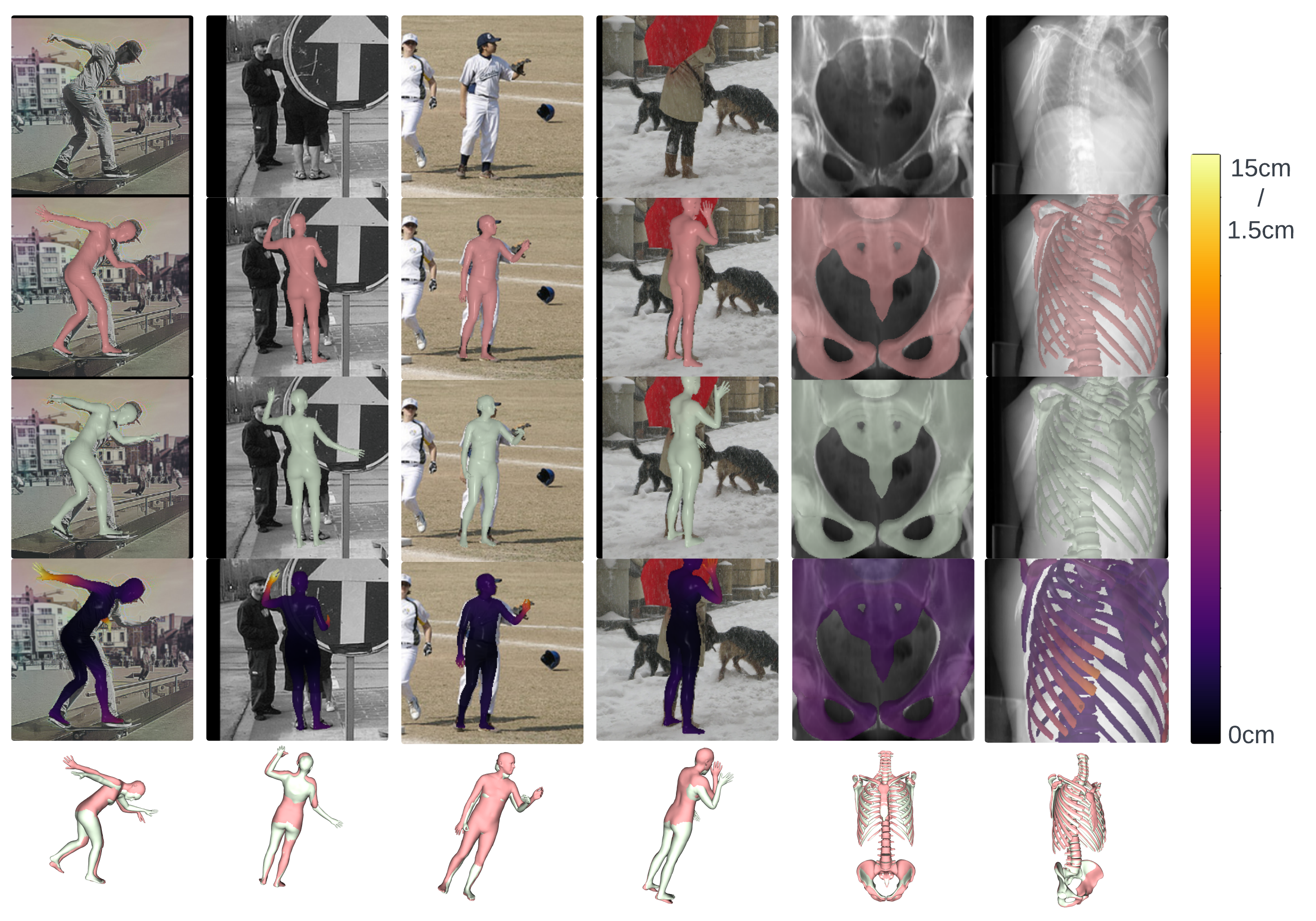}
 	\caption{	\label{fig:final} Qualitative results are displayed for both RGB and X-Ray images. Pink represents the mode of the distribution, while green shows a sample from the distribution. The fourth row quantifies the variance from the sampled poses in centimeters. The scale of variance is 15 cm for RGB images and 1.5 cm for X-Ray images. The last row provides a 3D depiction of the network's predictions. 
}
\end{figure*}

We further evaluate our approach in comparison to the original BEDLAM-CLIFF~\cite{bedlam} across different datasets. Notably, training exclusively on the AGORA~\cite{agora} dataset yields significant improvements, with even greater improvements observed when merging the AGORA and BEDLAM~\cite{bedlam} datasets.

During ablation for both scenarios, with and without the solver, we initially conduct training for 5 epochs focusing solely on reducing the pose NLL loss. Subsequently, training continues for 5 and 10 epochs with solver and without solver respectively, during which we apply 2D / 3D supervision.

\subsubsection{Why true $\text{SO}(3)$ distribution}
We develop a distribution comprising four rotations uniformly sampled, as shown in  Fig.~\ref{fig:figure1}. For visualization purposes, instead of showing the rotation directly, we represent it by rotating a vector along the z-axis. We train three Normalizing Flow models, which include GLOW~\cite{kolotouros2021probabilistic}, ReLie~\cite{sengupta2023humaniflow}, and our proposed model, all using a fixed context vector. Although this setup does not represent a real-world scenario, it sheds light on the characteristics of normalizing flow models in the context of rotations. Our observations reveal that the samples from GLOW are broadly dispersed, ReLie produces samples that are more sparsely distributed, and our proposed model using conditional Gaussian yields a distribution that aligns with the curvature which is significantly better way of approximating the distribution and sampling.

\section{Conclusion}
Our research introduces a novel probabilistic model for 3D human mesh recovery from 2D images, leveraging conditional normalizing flows to map inputs to distributions of plausible poses, facilitating diverse and accurate pose sampling. This model outperforms traditional single-estimate approaches and demonstrates superior performance on both single and multi-hypothesis on benchmarks like 3DPW. The introduction of Möbius transforms allows for a more accurate representation of 3D rotations on the $\text{SO}(3)$ manifold, thus refining the model's predictive capabilities. Our approach uniquely combines probabilistic pose sampling with deterministic keypoint prediction, enhancing 3D variation while also maintaining low 2D variations.  Our work not only showcases the model's efficacy in RGB imaging but also extends its application to X-Ray imagery, indicating its potential in medical scenarios. The integration of camera perspectives and shape variations into the probabilistic framework illustrates our model's adaptability and efficiency. Future directions could explore the application of this modeling technique to other object types and address ambiguities.

\paragraph*{Disclaimer}
The concepts and information presented in this article are based on research and are not commercially available.

\clearpage

\bibliographystyle{IEEEtran}
\bibliography{main}

\clearpage
\section*{Appendix}

\maketitle

\maketitle

\section{Introduction}
In this material, we detail the implementation and training aspects of our method, specifically focusing on the Möbius flow. We also delve into the datasets used and provide additional analysis of multiview results concerning RGB images. We also conduct experiments using the X-ray dataset and introduce a multi-view setup. Moreover, we offer additional qualitative results to demonstrate the performance of ProPLIKS, while also examining instances of failure cases.

\subsection{Datasets}\label{datasets}
\noindent\textbf{Human3.6M:} Human3.6M~\cite{h36m} is a key dataset for 3D human pose estimation, taken in indoor settings and offering multiple camera views. We use subjects S9 and S11 for testing, following Protocol 2 which means we focus on images from the front cameras. When we evaluate on multi-view setup, we use all four cameras available. We evaluate on the original joints (on 14 common LSP joints~\cite{kolotouros2019learning}), provided from the dataset.

\noindent\textbf{3DPW:} The 3DPW dataset~\cite{3dpw} is used for outdoor 3D pose and shape estimation. The SMPL parameters are recovered from IMU sensors.  We use 3DPW only for testing (on 14 common LSP joints~\cite{kolotouros2019learning}) our model  on the predefined test split to see how well it works in real-world conditions. We also perform ablations using this dataset.

\noindent\textbf{AGORA:} AGORA~\cite{agora} is a synthetic dataset with accurate SMPL models fitted to 3D scans, featuring approximately 85k image crops. We use this primarily for training the model.

\noindent\textbf{BEDLAM:} Like AGORA, BEDLAM~\cite{bedlam} is also a synthetic dataset. We only use the training part of BEDLAM dataset, which makes up around 750k image crops, to help train our model.

\subsection{Network Details}
\subsubsection{Encoder}
Our network architecture employs HRNet-48~\cite{hrnet} as its encoder backbone, adopting an encoder feature map akin to those utilized in PLIKS~\cite{shetty2023pliks} and PARE~\cite{kocabas2021pare}. The features extracted from this backbone produce feature maps $F \in \mathbb{R}^{780, c, c}$ along with a context feature vector $\mathbf{x} \in \mathbb{R}^{1024}$. Given an input image resolution of $224$ pixels, the dimensions of the feature maps' height and width would be $c = 56$.

\subsubsection{Pixel-Aligned Vertices}
The SMPL~\cite{smpl} and BOSS~\cite{boss} models contain approximately 6.9k and 65k vertices, respectively. Predicting all these vertices through a network would be memory intensive. Therefore, we opt to sample a total of $N_s < 1k$ vertices uniformly across different body regions. For both SMPL and BOSS models, each mesh vertex from the template is influenced by the kinematic chain, guided by blending weights. We classify the vertices into different body regions when their blending contribution exceeds 0.5, as depicted in Fig 2. from ~\cite{shetty2023pliks}. From these individual regions, we sample approximately 20 to 30 vertices using the farthest point sampling method. The number of vertices sampled from each segment is proportional to the segment's surface area—larger surfaces have more vertices sampled, while smaller surfaces have fewer. The indices of the sampled vertices are stored and utilized for all sampling purposes.

To adjust the feature map $F$'s channels from $780$ to $N$, we employ a single convolution layer. This is followed by flattening the feature map along its height and width dimensions. Subsequently, we apply two MLPs (Multi-Layer Perceptrons) to reduce the dimensionality from $c \times c$ to a single channel. This process is repeated four times to generate values for $\vecb{i} {=} $($\vecb{u}$, $\vecb{v}$), $\vecb{w}_v$ and $\vecb{w}_w$, which correspond to the pixel coordinates, variance factor, and weighting factor respectively.

\subsubsection{Global Rotaion}
In line with the approach taken by~\cite{sengupta2023humaniflow} , we utilize the context vector $\vecb{x}$ to ascertain the global body rotation. For ablation studies, where the network is trained without utilizing a solver, we also predict the camera parameters.

\subsubsection{Möbius Flow}
We provide a detailed explanation of the forward process, volume element, and inverse process of the Möbius Flow in Algorithm~\ref{alg:mf}, Algorithm~\ref{alg:mv}, and Algorithm~\ref{alg:mi}, respectively. The forward process is straightforward and elaborated on in the main paper. The variation in the volume element during the forward process, transitioning from $u_1$ to $\hat{u}_1$, is denoted by $\log(\left|\frac{\partial \theta}{\partial c_z}\right|)$, which reflects the Jacobian's determinant. Since altering the basis does not affect the volume within that space, we can straightforwardly determine the overall determinant of Jacobian using $\frac{\partial \theta}{\partial c_z}$. Our implementation's accuracy is confirmed through the use of autograd and Sympy for Jacobian computation.  For the inverse process, we employ a bisection search~\cite{rezende2020normalizing}, as this operation is monotonic. It involves adjusting the basis for $\omega$ until $\phi+\theta \approx 0$. This adjustment typically reaches the desired precision of $\epsilon = 10^{-4}$ within 18 iterations. 

Our model alternates between a set of Möbius coupling (specifically, two blocks, each representing two vectors from a rotation matrix) and rotation flow. In total, the architecture comprises four sets of Möbius coupling and four rotation flows, with Möbius layer having combination of 16 Möbius transformations. The transformations for both Möbius and rotation flows are provide with a residual neural network~\cite{kolotouros2021probabilistic}, with 256 channels in each residual block, and total of 2 blocks.

\begin{algorithm}
\caption{Möbius Flow Forward}\label{alg:mf}
\begin{algorithmic}
\Require $u_1, u_2, \omega, w$ 
\State $u_3 \gets u_1 \times u_2$
\State $c_z \gets \sum(u_1 \cdot u_1) + i \cdot \sum(u_1 \cdot u_3)$
\State $c_\omega \gets \sum(\omega \cdot u_1) + i \cdot \sum(\omega \cdot u_3)$\Comment{$\omega$ of shape $k\times3$}
\State $z_1 \gets \frac{c_z + c_\omega}{c_\omega^* \cdot c_z + 1}$
\State $\vecb{\theta} \gets \angle(z_1)$\Comment{angle of $z_1$}
\State $\theta \gets \sum(\vecb{\theta} \cdot w)$\Comment{Convex combination of k $\vecb{\theta}$}
\State $z_{\text{out}} \gets e^{i \cdot \theta}$
\State $\hat{u}_1 \gets \text{Re}(z_{\text{out}}) \cdot u_2 + \text{Im}(z_{\text{out}}) \cdot u_3$
\end{algorithmic}
\end{algorithm}

\begin{algorithm}
\caption{Möbius Flow Volume Element}\label{alg:mv}
\begin{algorithmic}
\State $\frac{\partial z_1}{\partial c_z} \gets  \frac{i \cdot (c_\omega + c_z) \cdot c_\omega^*}{(c_z \cdot c_\omega^* + 1)^2} - \frac{i}{(c_z \cdot c_\omega^* + 1)}$
\State $\frac{\partial \vecb{\theta}}{\partial c_z} \gets \frac{(-\text{Re}(z_1) \cdot \text{Im}(\frac{\partial z_1}{\partial c_z}) + \text{Im}(z_1) \cdot \text{Re}(\frac{\partial z_1}{\partial c_z}))}{\text{Re}(z_1)^2 + \text{Im}(z_1)^2}$
\State $\frac{\partial \theta}{\partial c_z} \gets \sum(\frac{\partial \vecb{\theta}}{\partial c_z} \cdot w)$
\State $\Delta V \gets \log(\left|\frac{\partial \theta}{\partial c_z}\right|)$
\end{algorithmic}
\end{algorithm}

\begin{minipage}{0.46\textwidth}
\begin{algorithm}[H]
\caption{Möbius Flow Inverse}\label{alg:mi}
\begin{algorithmic}
\Require $\hat{u}_1, u_2, \omega, w$
\State $u_3 \gets \hat{u}_1 \times u_2$
\State $\epsilon \gets 10^{-4}$
\State $L \gets -\pi , U \gets \pi $

\While {true}
    \State $M \gets (L + U) / 2$
    \State $\tau \gets \Call{Reverse}{M, \hat{u}_1, u_3, \omega, w}$
    \State $C \gets (\tau + M < 0)$
    \State $L \gets C ~ ?  ~ M : L$
    \State $U \gets C ~ ? ~ U : M$
    \State $\text{stop} \gets (\left|\tau + M\right| < \epsilon)$
    \If{$(\text{stop})$}
        \State \textbf{break}
    \EndIf
\EndWhile

\State $u_1 \gets \hat{u}_1 \cos(M) + u_3 \sin(M)$
\end{algorithmic}
\end{algorithm}
\end{minipage}
\hfill
\begin{minipage}{0.46\textwidth}
\begin{algorithm}[H]
\begin{algorithmic}
\Function{Reverse}{$M, \hat{u}_1, u_2, \omega, w$}
\State $c_M \gets e^{-i \cdot M}$
\State $c_z \gets 1 + 0\cdot i$
\State $c_\omega \gets (\sum(\omega \cdot \hat{u}_1) + i \cdot \sum(\omega \cdot u_2)) \cdot c_M$
\State $z_1 \gets \frac{c_z + c_\omega}{c_\omega^* \cdot c_z + 1}$
\State $\vecb{\theta} \gets \angle(z_1)$
\State $\theta \gets \sum(\vecb{\theta} \cdot w)$
\State \Return $\theta$
\EndFunction
\end{algorithmic}
\end{algorithm}

\end{minipage}

\subsection{Network Training}
Given that the entire pipeline is differentiable, we train the network in an end-to-end manner, dividing the training process into pre-training and subsequent training phases (incorporating both 2D and 3D supervision) to accelerate the training speed. During pre-training, we focus exclusively on training the pose, shape, and pixel-aligned vertices. However, for ablation studies without a solver (applicable to both ProPLIKS and ProHMR), pixel-aligned vertices are excluded. The loss objective minimized during pre-training is defined as follows:

\begin{equation}\label{eq_obj}
\mathcal{L} = \lambda_1 L_{\text{lnll}} + \lambda_2 L_{\theta\text{nll}} + \lambda_3 L_{\beta\text{nll}} + \lambda_4 L_{\theta_g} .
\end{equation}

Here, $L_{\theta\text{nll}} = -\ln p_{\Theta \mid I}(\vec{\theta}{\text{gt}} \mid \mathbf{x})$ aims to optimize the negative log likelihood of a given ground truth (GT) pose conditioned on the image, focusing on the body pose excluding the global pose. $L_{\text{lnll}}$ represents the Laplacian negative log likelihood for the pixel-aligned vertices, $L_{\beta\text{nll}}$ is analogous to $L_{\theta\text{nll}}$ but for shape parameters, and $L_{\theta_g}$ quantifies the L2 distance between the predicted and GT global rotation of the subject.

Following the pre-training phase, we introduce further supervision for both 2D ($\lambda_5L_{2D}$) and 3D ($\lambda_6L_{3D}$) joint positions, aiming to minimize the L1 distance between the predicted joints and their corresponding ground truth. Notably, the 3D loss is applied solely to the mode of the distribution. This approach enables the network to capture a wider variety of 3D pose variations.

This two-step training process is implemented owing to the inclusion of the PLIKS solver and to mitigate ambiguous rotations during the initial training. In the initial stages of training, we would notice numerical instability; specifically, the predictions of pixel-aligned vertices lack the consistency needed for effective solver application, leading to ill-posed reconstructions. Furthermore, the intensity of the 3D loss may lead to inaccuracies in pose prediction despite correct 3D joint positioning, as joints are represented as directionless points in 3D space.

\subsection{Implementation Details}
PyTorch~\cite{pytorch} is used for implementation. For all our experiments we initialize the HRNet~\cite{hrnet} backbone with weights pre-trained on the COCO~\cite{coco} dataset, which exhibits faster convergence during training. We use the Adam optimizer~\cite{adam} with a mini-batch size of 64. We split the training into 2 steps, training the normalizing flow, followed by training with 2D and 3D supervision loss. We follow the same for ablations too. The learning rate at pre-training is set to $1e^{-4}$, whereas, while training the entire pipeline it is initialized to $1e^{-5}$. The network is pre-trained for 3 epochs, and then finally trained for further 5 epochs. For ablations, with no solver, the network is pre-trained for 3 epochs followed by 10 epochs with supervision. It takes less than a day to train on a single NVIDIA A100 GPU. We set $\lambda_1$, $\lambda_2$, $\lambda_3$, $\lambda_4$, $\lambda_5$,, and $\lambda_6$ to 0.01, 0.001, 0.0001, 1, 0.1,  and 0.1, respectively.

For the X-ray experiments we obtain ground truth data by fitting 

\section{Experiments}
We present additional experiments on X-ray images and introduce a multi-view setup, where we utilize the network trained on single views without retraining. We also demonstrate multi-view results on the Human3.6M dataset.

\subsection{X-Ray Images}

In the case of X-Ray images, our training utilized the CTLymph~\cite{lymph} and AutoPet~\cite{autopet} datasets. For validation, we utilized the QIN-HeadNeck~\cite{qin} and Visceral~\cite{visceral} datasets.  We created Digitally Reconstructed Radiograph (DRR) images from CT scans as synthetic inputs for the network. and applied the BOSS model to segmentations provided by TotalSegmentator~\cite{wasserthal2023totalsegmentator}.

The process of obtaining BOSS ground truth fits involves converting segmented CT scans into surface meshes. We use TotalSegmentator~\cite{wasserthal2023totalsegmentator} to segment the bone structures and then transform these segmentations into surface meshes. Traditional iterative optimization techniques are then applied to fit the BOSS model to these surface scans. The parameters obtained from this process are used for both training and validation purposes.

Our model's performance was assessed using digitally reconstructed radiographs (DRRs) from the X-Ray test set. Quantitative results for pose estimation are detailed in Table~\ref{tab:xray1}. In contrast to RGB images, patients in X-Ray images are typically positioned similarly, with most variations arising from differences in body shape and camera angles. Rather than capturing the entire bone structure, it's more common to obtain images of smaller, specific sections. The evaluation metrics are divided into three areas: pelvis, lumbar, and thorax. Unlike the RGB images where joint-based metrics (MPJPE) are used, here we evaluate based on keypoints defined on the mesh surface. This approach accounts for the complexity of human bone structure, making the evaluation more robust by selecting vertices. The number of keypoints is determined by each bone segment's surface area, with a minimum of four per segment. We conducted a comparison with our own implementation of ProHMR~\cite{kolotouros2021probabilistic}, utilizing the same backbone and training strategies as detailed in Table~\ref{tab:xray1}. For the relatively simpler and larger pelvis structure, we noted an error of 11.10 mm in the visible region. The lumbar region, being more intricate, showed a slightly increased error of 22.09 mm. The thorax, encompassing complex structures like ribs, vertebrae, scapula, and clavicle, exhibited the highest error rate of 33.66 mm. 

\begin{table*}[t]
\centering
\begin{tabular}{l|cccc}
       & \multicolumn{2}{c}{All}                          & \multicolumn{2}{c}{Visible} \\ \cline{2-5} 
       & MPJPE$\downarrow$~(mm)        & \multicolumn{1}{c|}{PA-MPJPE$\downarrow$~(mm)}     & MPJPE$\downarrow$~(mm)        & 2DKP$\downarrow$~(px)     \\ \cline{1-5} 
Pelvis \textbf{[Proposed]} & 32.56~(24.97) & \multicolumn{1}{c|}{17.28~(15.30)} & 11.10~(8.87)  & 7.34~(7.78)   \\ \myrowcolour
Lumbar \textbf{[Proposed]} & 29.71~(22.52) & \multicolumn{1}{c|}{16.41~(14.63)} & 22.09~(16.30) & 11.82~(14.35) \\
Thorax \textbf{[Proposed]} & 32.09~(24.61) & \multicolumn{1}{c|}{15.91~(14.31)} & 33.66~(24.83) & 9.37~(10.53) \\   \hline\myrowcolour 
Pelvis~\cite{kolotouros2021probabilistic} & 35.03~(26.56) & \multicolumn{1}{c|}{17.30~(16.08)} & 12.53~(9.86)  & 9.23~(10.65)   \\ 
Lumbar~\cite{kolotouros2021probabilistic} & 32.57~(26.02) & \multicolumn{1}{c|}{16.22~(15.22)} & 22.68~(17.88)  & 17.00~(18.23)   \\ \myrowcolour
Thorax~\cite{kolotouros2021probabilistic} & 36.09~(28.07) & \multicolumn{1}{c|}{16.71~(15.79)} & 37.28~(27.87)  & 15.91~(16.42)   
\end{tabular}
\caption{\label{tab:xray1} Assessment on a single-image X-Ray dataset is presented. The values in brackets indicate the minimum for 3D metrics (mean for 2D metrics) from a sample size of $n = 50$, while the other value denotes the mode. Here, Visible refers to keypoints that are detectable within the imaging field.}
\end{table*}

\subsection{RGB Multi-View}

\begin{figure*}[ht]
	\centering
	\begin{subfigure}[b]{0.49\textwidth}
	\centering
	    \includegraphics[trim={1cm 7cm 1cm 7cm },clip,frame, width=\textwidth,keepaspectratio]{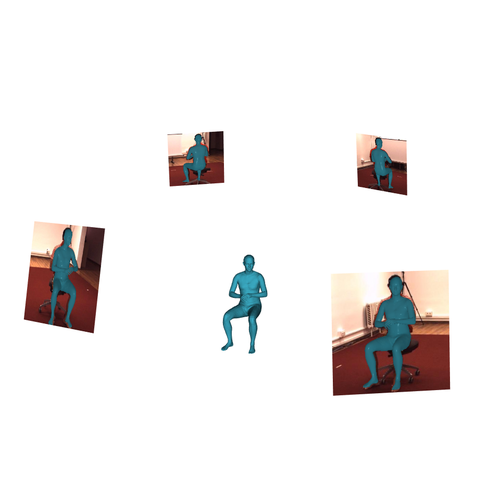}
         \includegraphics[trim={1cm 7cm 1cm 7cm},clip,frame,width=\textwidth,keepaspectratio]{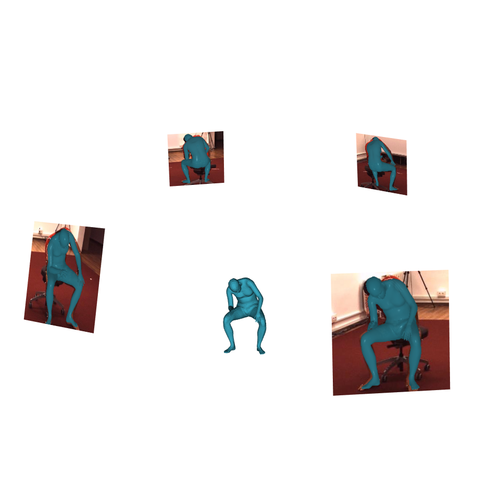}
    \end{subfigure}
    \begin{subfigure}[b]{0.49\textwidth}
	\centering
	    \includegraphics[trim={1cm 7cm 1cm 7cm },clip,frame,width=\textwidth,keepaspectratio]{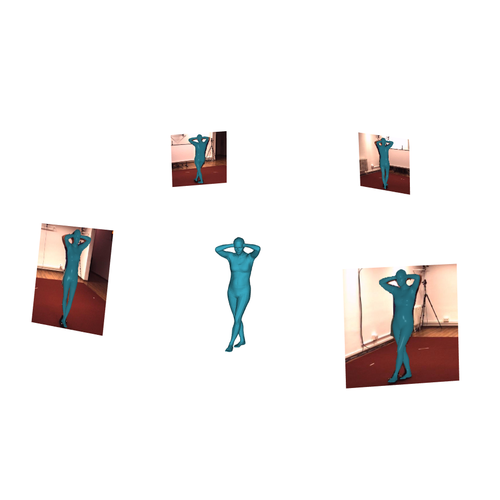}
         \includegraphics[trim={1cm 7cm 1cm 7cm},clip,frame,width=\textwidth,keepaspectratio]{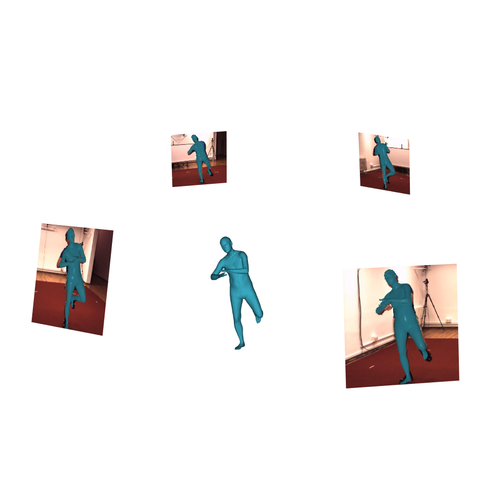}
    \end{subfigure}
 	\caption{	\label{fig:mv} Qualitative results on multiview images (from 4 cameras) on the  Human3.6m~\cite{h36m} dataset.}
\end{figure*}

Using the same trained model, we perform multiview reconstruction on the Human3.6m dataset employing 1, 2, and 4 cameras. As indicated in Table~\ref{tab:mv}, the employment of more than one camera enhances the performance in terms of both MPJPE and absolute MPJPE errors. Conversely, the 2DKP error exhibits variability due to the optimization process, which is tasked with optimizing for 2D views across multiple perspectives. Qualitative outcomes are showcased in Fig.~\ref{fig:mv}.

\begin{table*}[]
\centering
\begin{tabular}{c|c|c|c|c}
    & MPJPE$\downarrow$~(mm) & MPJPE (abs)$\downarrow$~(mm) & PA-MPJPE$\downarrow$~(mm) & 2DKP$\downarrow$~(px) \\ \hline
N=1 & 73.6                   & 170.5                        & 44.4                     & 6.58                   \\ \myrowcolour
N=2 & 63.9                   & 66.2                         & 38.1                      & 6.02                  \\
\textbf{N=4} & 60.6                   & 55.6                         & 34.8                      & 5.50                 
\end{tabular}
\caption{\label{tab:mv} Assessment on a multi-view Human3.6m dataset is presented. Here, abs refers absolute distance, while N=1 referes to single view, wheras N$>$1 refers multi-view.}
\end{table*}

\section{Qualitative Results}

In this section, we provide additional qualitative results.

\subsection{Visualizing Uncertainty}
\begin{figure*}[ht]
\centering
\begin{subfigure}[b]{0.15\textwidth}
    \centering
    \includegraphics[width=\textwidth,keepaspectratio] {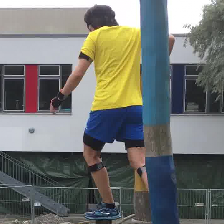}
    \includegraphics[width=\textwidth,keepaspectratio] {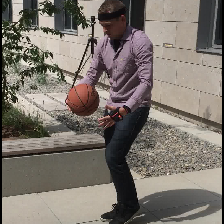}
    \includegraphics[width=\textwidth,keepaspectratio] {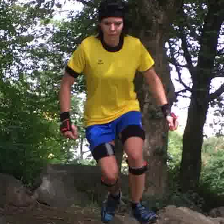}

\end{subfigure}
\begin{subfigure}[b]{0.15\textwidth}
    \centering
    \includegraphics[width=\textwidth,keepaspectratio] {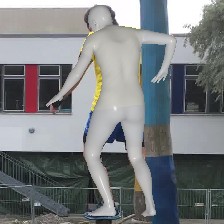}
    \includegraphics[width=\textwidth,keepaspectratio] {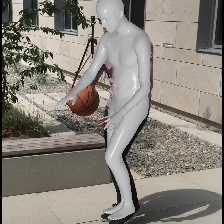}
    \includegraphics[width=\textwidth,keepaspectratio] {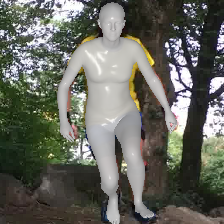}
\end{subfigure}
\begin{subfigure}[b]{0.15\textwidth}
    \centering
    \includegraphics[width=\textwidth,keepaspectratio] {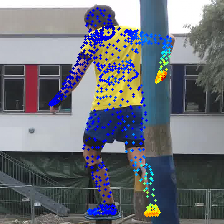}
    \includegraphics[width=\textwidth,keepaspectratio] {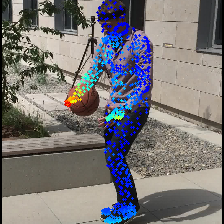}
    \includegraphics[width=\textwidth,keepaspectratio] {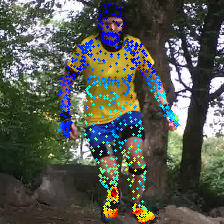}
\end{subfigure}
\begin{subfigure}[b]{0.15\textwidth}
    \centering
    \includegraphics[width=\textwidth,keepaspectratio] {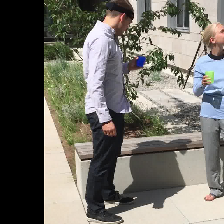}
    \includegraphics[width=\textwidth,keepaspectratio] {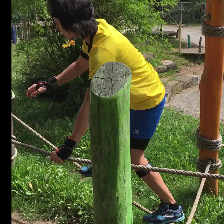}
    \includegraphics[width=\textwidth,keepaspectratio] {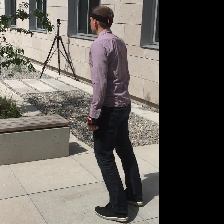}
\end{subfigure}
\begin{subfigure}[b]{0.15\textwidth}
    \centering
    \includegraphics[width=\textwidth,keepaspectratio] {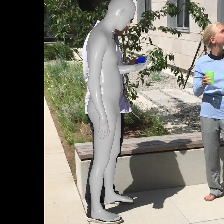}
    \includegraphics[width=\textwidth,keepaspectratio] {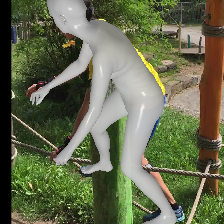}
    \includegraphics[width=\textwidth,keepaspectratio] {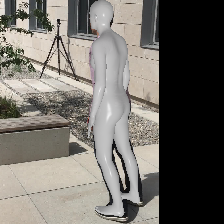}
\end{subfigure}
\begin{subfigure}[b]{0.15\textwidth}
    \centering
    \includegraphics[width=\textwidth,keepaspectratio] {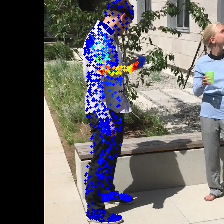}
    \includegraphics[width=\textwidth,keepaspectratio] {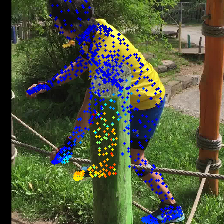}
    \includegraphics[width=\textwidth,keepaspectratio] {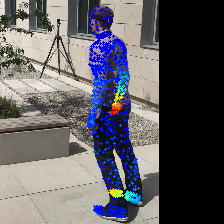}
\end{subfigure}
\caption{	\label{fig:unc} Visualization of uncertainty in pixel aligned vertices. Here, red cross indicates highly uncertain vertex predictions whereas blue indicates vertices with confidence. The scale has been normalized per image.}
\end{figure*}

Several instances of vertex uncertainty predictions are presented in Fig.~\ref{fig:unc}. These illustrations show that in situations involving occlusions or uncertain depths, the vertices exhibit increased variance. When these predictions are input into the solver as weights, it results in greater flexibility for adjustments in poses within these regions.

\subsection{Failure Mode}
\begin{figure*}[h]
\centering
\begin{subfigure}[b]{0.15\textwidth}
    \centering
    \includegraphics[width=\textwidth,keepaspectratio] {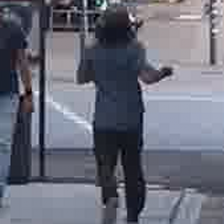}
    \includegraphics[width=\textwidth,keepaspectratio] {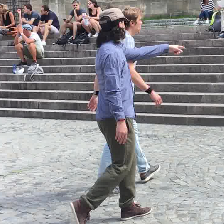}
\end{subfigure}
\begin{subfigure}[b]{0.15\textwidth}
    \centering
    \includegraphics[width=\textwidth,keepaspectratio] {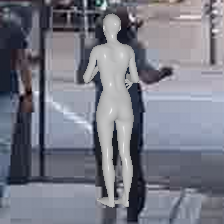}
    \includegraphics[width=\textwidth,keepaspectratio] {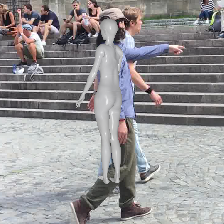}
\end{subfigure}
\begin{subfigure}[b]{0.15\textwidth}
    \centering
    \includegraphics[width=\textwidth,keepaspectratio] {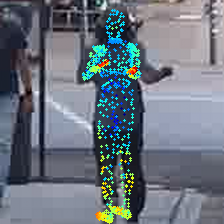}
    \includegraphics[width=\textwidth,keepaspectratio] {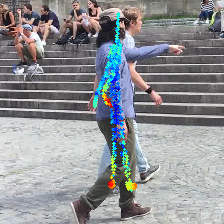}
\end{subfigure}
\begin{subfigure}[b]{0.15\textwidth}
    \centering
    \includegraphics[width=\textwidth,keepaspectratio] {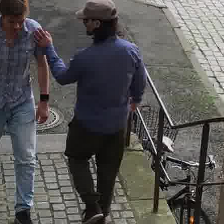}
    \includegraphics[width=\textwidth,keepaspectratio] {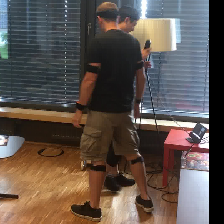}
\end{subfigure}
\begin{subfigure}[b]{0.15\textwidth}
    \centering
    \includegraphics[width=\textwidth,keepaspectratio] {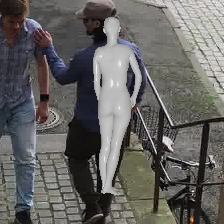}
    \includegraphics[width=\textwidth,keepaspectratio] {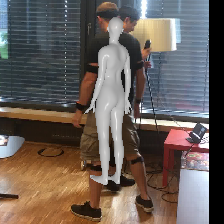}
\end{subfigure}
\begin{subfigure}[b]{0.15\textwidth}
    \centering
    \includegraphics[width=\textwidth,keepaspectratio] {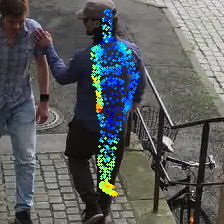}
    \includegraphics[width=\textwidth,keepaspectratio] {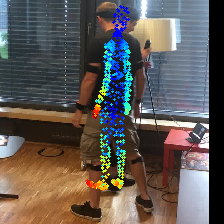}
\end{subfigure}
\caption{	\label{fig:fail} Displaying examples of failure scenarios, which may stem from occlusion, the presence of multiple people, or inaccurate vertex predictions.}
\end{figure*}

In Fig.~\ref{fig:fail}, we present several instances where our technique does not successfully reconstruct plausible human body positions. These failures can be attributed to scenarios with multiple individuals within the frame, occlusions, or erroneous vertex predictions. Given that our approach to vertex prediction relies on a straightforward CNN followed by two MLPs, enhancing this aspect could serve as a potential avenue for future improvements.

\subsection{Qualitative}
\begin{figure*}[h]
\centering
\begin{subfigure}[b]{0.15\textwidth}
    \centering
    \includegraphics[width=\textwidth,keepaspectratio] {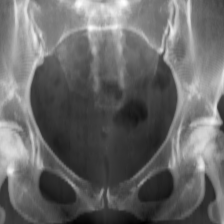}
    \includegraphics[width=\textwidth,keepaspectratio] {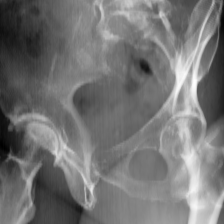}
\end{subfigure}
\begin{subfigure}[b]{0.15\textwidth}
    \centering
    \includegraphics[width=\textwidth,keepaspectratio] {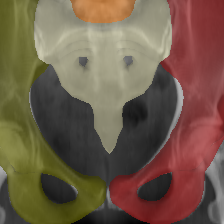}
    \includegraphics[width=\textwidth,keepaspectratio] {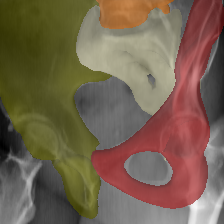}
\end{subfigure}
\begin{subfigure}[b]{0.15\textwidth}
    \centering
    \includegraphics[width=\textwidth,keepaspectratio] {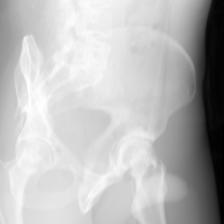}
    \includegraphics[width=\textwidth,keepaspectratio] {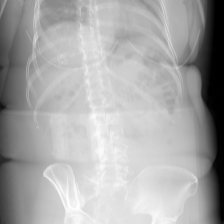}
\end{subfigure}
\begin{subfigure}[b]{0.15\textwidth}
    \centering
    \includegraphics[width=\textwidth,keepaspectratio] {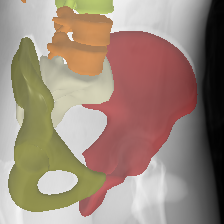}
    \includegraphics[width=\textwidth,keepaspectratio] {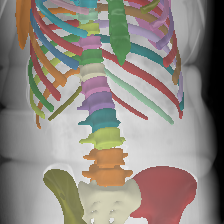}
\end{subfigure}
\begin{subfigure}[b]{0.15\textwidth}
    \centering
    \includegraphics[width=\textwidth,keepaspectratio] {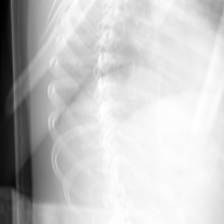}
    \includegraphics[width=\textwidth,keepaspectratio] {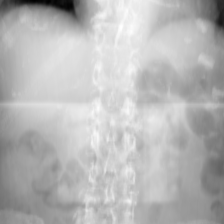}
\end{subfigure}
\begin{subfigure}[b]{0.15\textwidth}
    \centering
    \includegraphics[width=\textwidth,keepaspectratio] {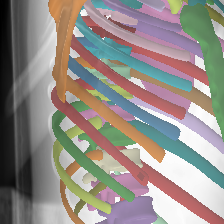}
    \includegraphics[width=\textwidth,keepaspectratio] {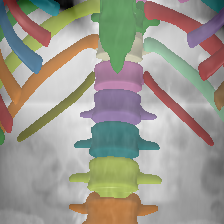}
\end{subfigure}
\caption{	\label{fig:xray} Examples on X-ray images.}
\end{figure*}

\begin{figure*}[h]
\centering
\includegraphics[width=0.9\textwidth,keepaspectratio]{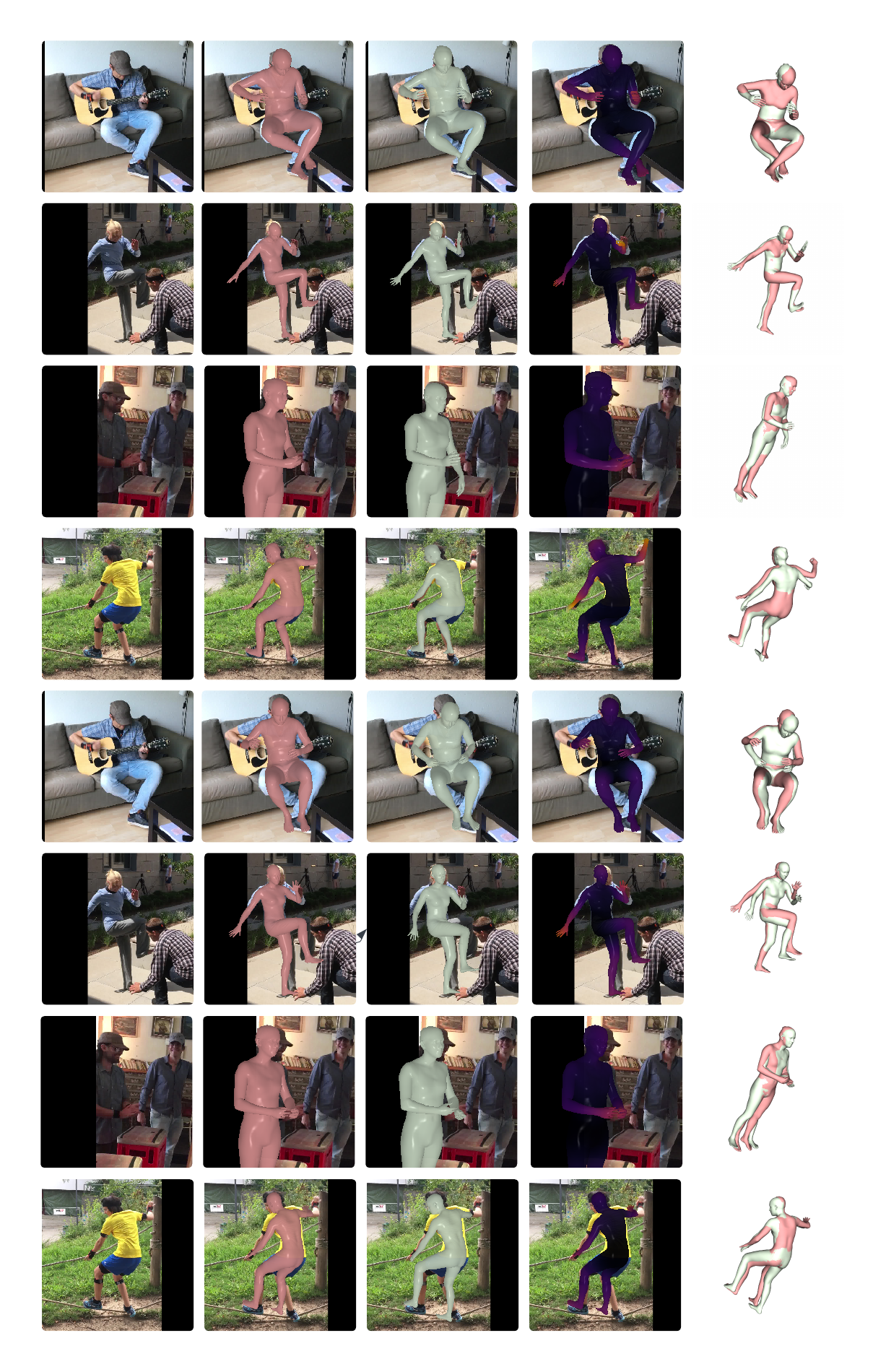}
\vspace{-18pt}
\caption{	\label{fig:final2} Qualitative results on multi-hypothesis from random samples on PW3D dataset~\cite{3dpw}. Top four rows are based on the proposed solution, bottom four are based on GLOW~\cite{kolotouros2021probabilistic}.}
\end{figure*}

In Fig.~\ref{fig:xray} and Fig.~\ref{fig:final2}, we provide additional qualitative results on X-ray and RGB images.

\end{document}